\documentclass[review]{elsarticle}

\usepackage{float}
\usepackage{multirow}
\usepackage{lineno,hyperref}
\modulolinenumbers[5]
\usepackage{amsmath}
\usepackage{algorithm}
\usepackage{algpseudocode}

\journal{Indian Institute of Technology - Kanpur, India}

\begin{document}

\begin{frontmatter}

\title{Serpentine Synergy: Design and Fabrication of a Dual Soft Continuum Manipulator and Soft Snake Robot}

\author{Rajashekhar V S \fnref{Corresponding author}}
\address{Research Scholar, Department of Design, Indian Institute of Technology - Kanpur, Uttar Pradesh, India}
\fntext[myfootnote]{Corresponding author}
\ead{raja23@iitk.ac.in}

\author{Aravinth Rajesh}
\address{Student, National Institute of Technology - Tiruchirappalli, Tamil Nadu, India}
\ead{111121018@nitt.edu}

\author{Muhammad Imam Anugrahadi Athaaillah}
\address{Student, National Institute of Technology - Tiruchirappalli, Tamil Nadu, India}
\ead{111121074@nitt.edu}

\author{Gowdham Prabhakar}
\address{Assistant Professor, Department of Design, Indian Institute of Technology - Kanpur, Uttar Pradesh, India}
\ead{gowdhampg@iitk.ac.in}

\begin{abstract}
This work presents a soft continuum robot (SCR) that can be used as a soft continuum manipulator (SCM) and a soft snake robot (SSR). This is achieved using expanded polyethylene foam (EPE) modules as the soft material. In situations like post-earthquake search operations, these dual-purpose robots could play a vital role. The soft continuum manipulator with a camera attached to the tip can manually search for survivors in the debris. On the other hand, the soft snake robot can be made by attaching an active wheel to the soft continuum manipulator. This mobile robot can reach places humans cannot and gather information about survivors. This work presents the design, fabrication, and experimental validation of the dual soft continuum robot.     
\end{abstract}


\end{frontmatter}

\linenumbers

\section{Introduction}
The field of robotics is fascinating, especially when bio-inspiration is drawn from reptiles. These reptiles can traverse rough terrain, making them unique among other species of the world. Among them, snakes are versatile and exhibit several gaits based on the surface they traverse. Bio-mimicking them is a challenge, and several successful attempts have been made in the past \cite{wright2007design, rajashekhar2015serial}. It has been mentioned in \cite{chitikena2023robotics} that snake robots have a huge potential for being used for search and rescue (SAR) operations. The ethical and design considerations for snake robots that can be used for SAR operations have also been described. However, most snake robots in the literature are made of rigid mechanisms that can be used only as mobile robots. They cannot be used as a manipulator. This problem needs to be addressed since there are many situations like post-earthquake search operations (inside debris) and search for survivors inside collapsed buildings (due to several reasons), which would use the assistance of robots to aid humans in the process. Due to their sleek body design, the snake robots make it easy to pass through narrow gaps. It is to be noted that rigid-bodied snake robots have a limited number of degrees of freedom, whereas soft-bodied robots have infinite degrees of freedom \cite{luo2014theoretical,luo2015slithering} and can adapt to the path that they travel. 

The robots traversability through challenging terrains remains under extensive research due to the need for a universally optimal solution. This spans from traversability-estimation techniques to design implementations in the robot. The traversability-estimation techniques determine the robot's feasibility in moving through its surrounding terrain and are subdivided into three categories: conventional machine learning, deep learning, and non-trainable methods. Notably, the developing techniques from deep learning demonstrate superiority due to their ability to automatically learn complicated features, enabling them to handle more complex terrains with relative ease \cite{sevastopoulos2022survey}. Additionally, traversibility is not only dependent on the terrain but also on the robot's structure. More terra dynamically streamlined shapes, such as ellipsoids, are found to traverse through cluttered terrains with less effort \cite{li2015terradynamically}.


One of the difficulties with a wheeled robot is quick and precise maneuvering, for which inspiration can be drawn from animals like cheetahs \cite{machairas2015quadruped, briggs2012tails} and alligators \cite{willey2004tale}. These animals rely heavily on their tails for stabilization, which can be replicated on robots \cite{saab2019design}. The robotic tails can also be programmed to offer stability during fall \cite{chu2023combining}. The tail designs mentioned above offer aid while in dynamic motion but would provide limited assistance in active quasi-static navigation. These are when manipulator-like tails come into the picture, acting as limbs, which can interact with the terrain/environment and thus manipulate it. In this recent advancement in tail design, robots like Tail STAR (TSTAR) use tails to overcome obstacles, climb high steps (6 times its wheel radius), and bridge gaps equal to its body length \cite{coronel2023overcoming}. 

Despite recent progress on tail design, it's still limited in scope due to material and mass constraints. Exploring tails made for multiple tasks and liberating them beyond rigid structure is critical to improving tail utility \cite{saab2018robotic}. Most of the current search and rescue robots are complex robots as they have high structural integrity and accuracy. Still, they are made of hard materials with invariable properties, limited adaptability to operate in unknown environments, and fixed degrees of freedom \cite{alici2018softer}. Soft robotic tails permit embedding all the sensors within it and still provide infinite degrees of freedom \cite{russo2023continuum}. The wheeled robots' usage in rugged terrain can be further improved by integrating innovative systems like deep learning, computer vision, and machine learning for better traversability \cite{sevastopoulos2022survey}. To move in confined spaces, like during disaster aid, the robot should be able to manipulate its overall shape and size like cockroaches \cite{jayaram2016cockroaches} without compromising its speed and utility. 

On the other hand, the continuum robots inspired by snakes and elephant trunks have infinite degrees of freedom and thus provide high maneuverability, adaptability, and flexibility while navigating through their surrounding environment. These tails are categorized mainly as continuous backbone robots with a constant flexible body and segmented backbone robots with alternating flexible and rigid elements. The actuation for these robots can be done extrinsically, where the motion is generated at the base and transferred through the following links intrinsically, where the motion is generated locally at each backbone segment. The continuum robots exist between rigid-linked robots and the emerging soft-robots; a hybrid between these two models is referred to as soft-continuum manipulators \cite{russo2023continuum}.

\begin{figure}[H]
\begin{center}
\includegraphics[scale=0.45]{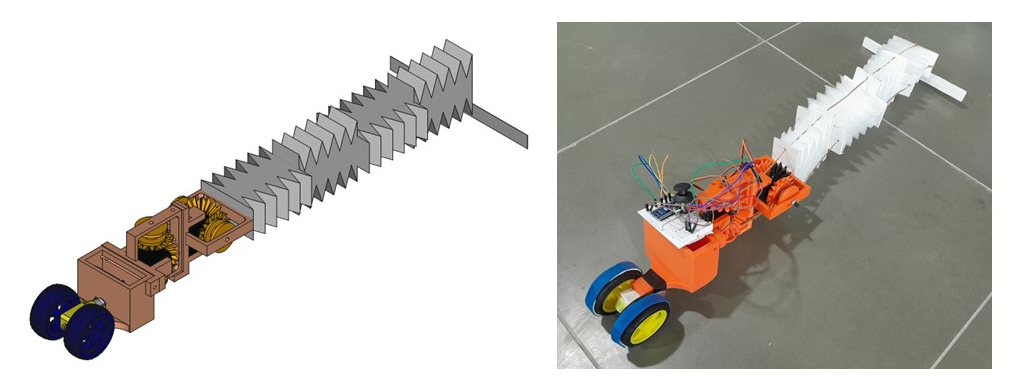}
\caption{(a) The computer aided design of the soft continuum robot (b) The fabricated model of the soft continuum robot}
\label{fig_intro}
\end{center}
\end{figure}

To overcome the problem of separate robots being used as a manipulator and mobile robots, dual soft continuum robots can be used \cite{rajashekhar2024developments}. It can be used in post-earthquake search operations. The soft continuum robot, in the form of soft continuum manipulators (SCM) fitted with a camera as an end effector, can be inserted inside the debris to search for survivors. If the void is big enough for the soft snake robot (SSR) to traverse, the same SCM can be fitted with a wheel and sent inside the cluttered environment to conduct the search operation. Using the soft continuum robot as a hand-held manipulator and mobile robot makes it a dual-purpose soft continuum robot. The computer-aided design can be seen in Figure \ref{fig_intro} (a). The fabricated robot can be seen in Figure \ref{fig_intro} (b). 

The rest of the paper is organized as follows: In Section \ref{sec_design}, the design of SCM has been explained. The optimal design of the module is presented in Section \ref{sec_optimal_design}. Section \ref{sec_fab} describes the fabrication of the modules and drive system used in the robot. Section \ref{sec_mechatronic_design} describes the electronics involved in the soft continuum robot (SCR). The working of the SCR is presented in Section \ref{sec_work}. The experimental results are presented in Section \ref{sec_experi}. The discussions are done in Section \ref{sec_discussion}. The limitations and future directions are presented in Section \ref{sec_lim_fut_dir}. Finally, the concluding remarks are given in Section \ref{sec_conc}. The appendix in Section \ref{append_Kinematic} explains the algorithm for kinematic analysis of the SSR. In Section \ref{append_workspace}, the algorithm for finding the theoretical workspace of the SCM is presented. Section \ref{append_geo_pro} uses the algorithm presented to find the module's thickness. Section \ref{append_grey} is used to find the height of the triangle for the fringes and the radius of curvature.

\section{Design of the Continuum Manipulator and Soft Snake Robot}
\label{sec_design}
The design of the soft continuum robot is presented in this section. The CAD is given in Section \ref{subsec_cad}. The design of the module and drive are provided in Section \ref{subsec_module} and \ref{subsec_drive_design}, respectively. The design of SCM and SSR are described in Section \ref{subsec_scm} and \ref{subsec_ssr}, respectively. The modeling of both modes is briefly explained in Section \ref{subsec_model}. The workspace of the SCM is calculated in Section \ref{subsec_workspace}. 
\subsection{Computer Aided Design of the Robot}
\label{subsec_cad}
\begin{figure}[H]
\begin{center}
\includegraphics[scale=0.32]{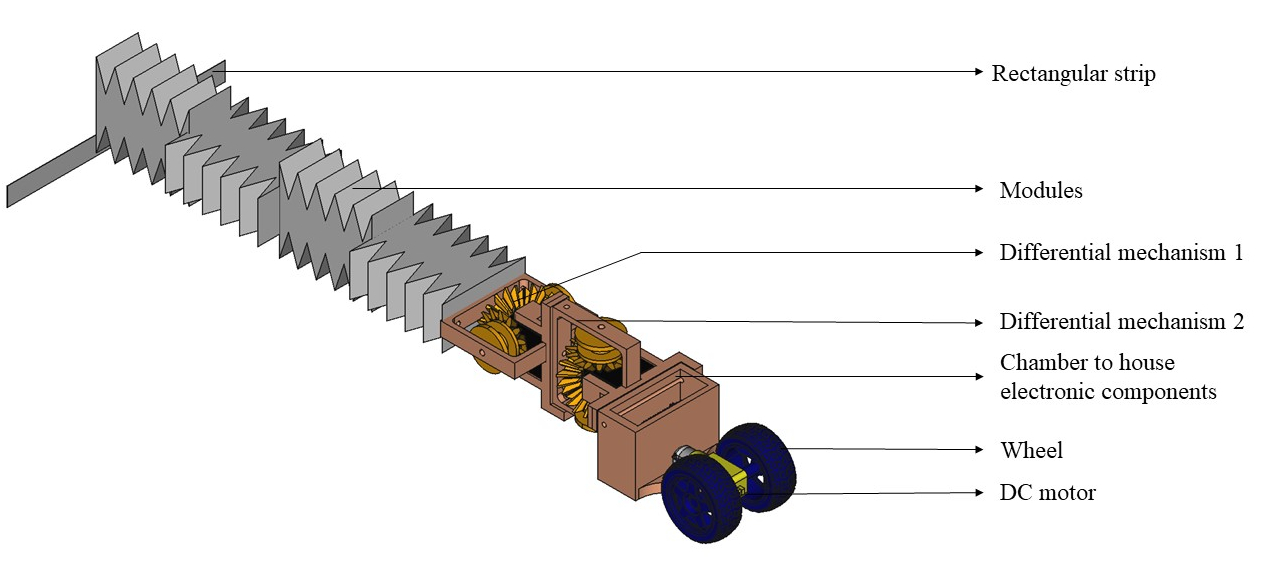}
\caption{The computer aided design of the soft continuum robot with the parts labelled}
\label{fig_CAD}
\end{center}
\end{figure}

The dual SCR was modeled using the \textit{FreeCAD$^{TM}$} software. It is shown in Figure \ref{fig_CAD}. A dual-shaft DC motor with a wheel attached to each side is placed in the front of the SSR. In the chamber behind the wheel, the electronic components of the robot are placed. Then, it is followed by the two differential drive systems, where the cables are wound. This is followed by the four soft modules placed in orthogonally offset position (in series). This is the vital part of the robot. The working of the cable-driven system is explained in Section \ref{subsec_drive_design}. The bevel gears for the dive system were created using the gear workbench available in the \textit{FreeCAD$^{TM}$} software. The various gear parameters, such as the number of teeth, module, and the thickness of the gear, were given as input. The bevel gears were generated accordingly by considering the other default parameters present in the workbench. Then, a pulley was attached to each driven bevel gear to wind the cable. A thin rectangular strip is connected to the end of the last module, which aids in turning the SSR. 

\subsection{Module Design}
\label{subsec_module}
\begin{figure}[H]
\begin{center}
\includegraphics[scale=0.15]{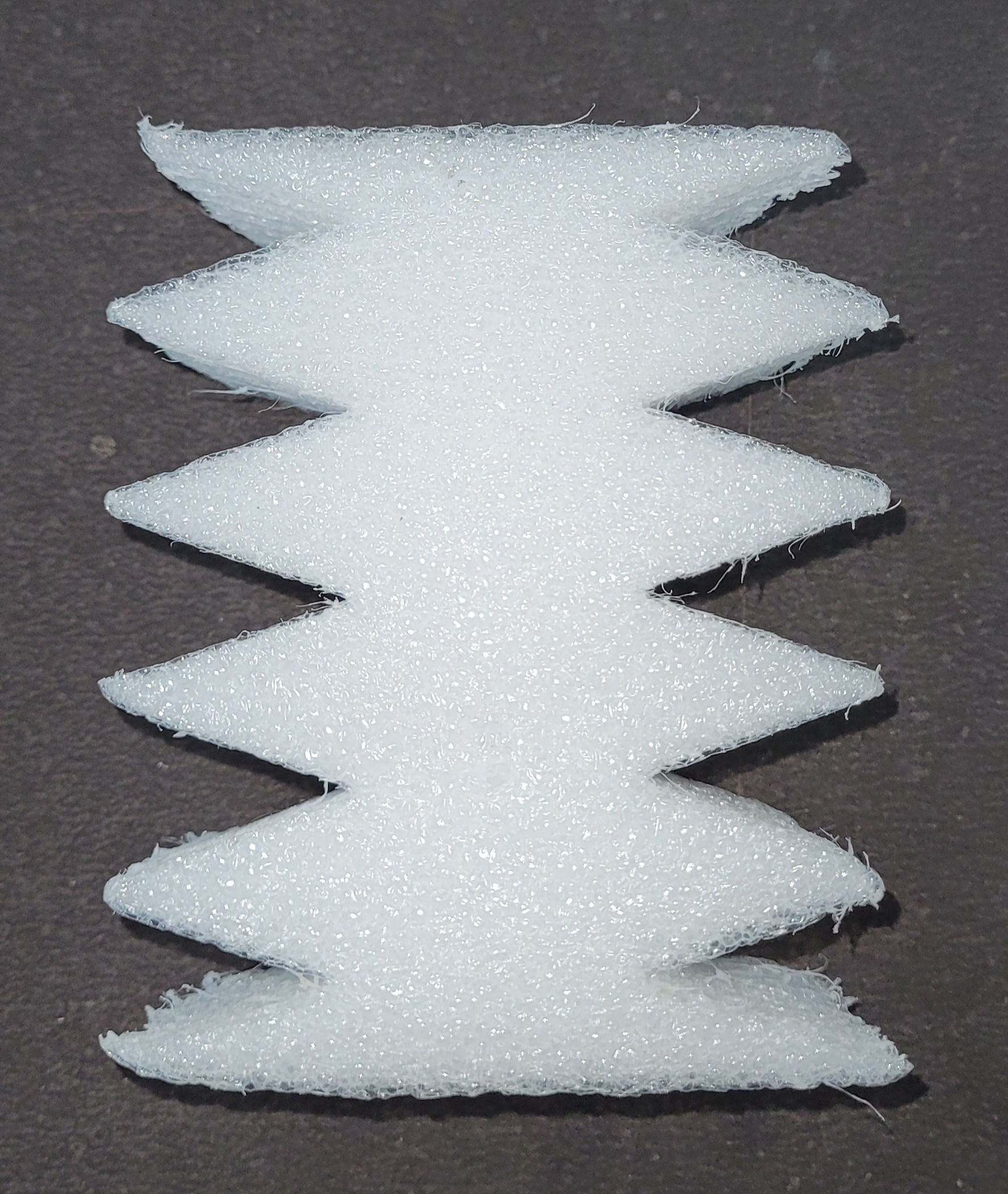}
\caption{The fabricated soft part that is used as the modules of the soft continuum robot}
\label{fig_module}
\end{center}
\end{figure}

\begin{figure}[H]
\begin{center}
\includegraphics[scale=0.4]{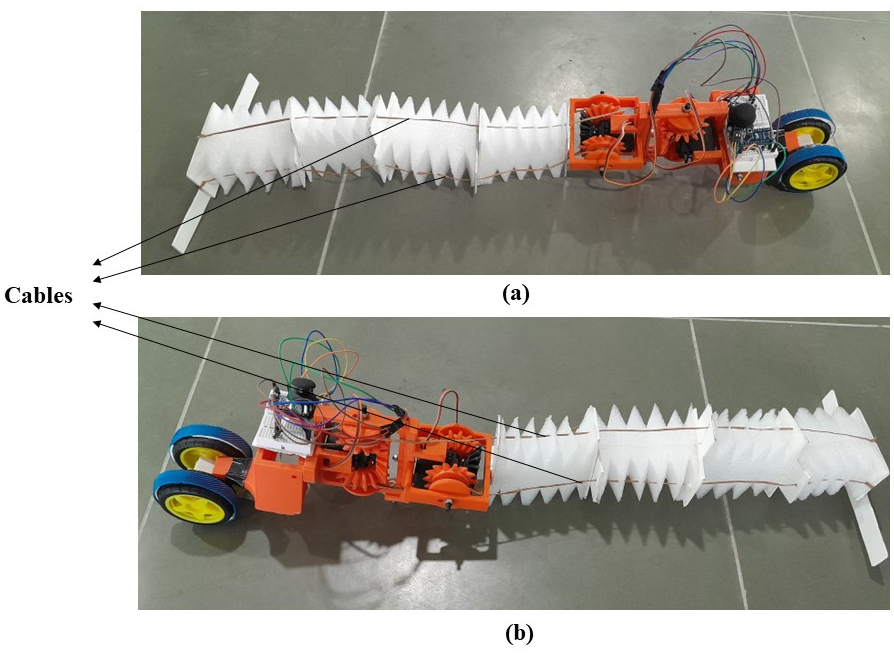}
\caption{The four cables that are used to drive the soft continuum robot (a) Left side view of the robot (b) Right side view of the robot}
\label{fig_drive}
\end{center}
\end{figure}

The cuboid module consists of a sawtooth structure on the sides, where they can compress or expand when they bend. There are four modules in the robot, each of which is capable of bending about an axis. The cross-section of the modules is chosen to be rectangular. The rectangular cross-sections provide greater directional flexibility and allow movement only along specific axes. The modules are attached in orthogonally offset directions in series so that they bend about their axis in the 3D space. A module that is used in the robot is shown in Figure \ref{fig_module}. Four cables run through the modules as shown in Figure \ref{fig_drive}. They help bend the modules when pulled with force from one end. The other end of the cables is fixed to the last module. The drive system takes care of how the cables are actuated. The advantage of the cable-driven mechanism is that it can provide faster response, long-distance force transmission, high precision, and is lightweight, and compact \cite{zhou2022bioinspired}. It has a disadvantage that the friction between the cables and modules can affect the SCMs modeling and control.

\subsubsection{Calculating the number of Fringes}
\label{subsub_fringes}
The module is made up of fringes at the sides. To calculate the number of fringes (N), the base (b) and height (h) of each triangular fringe need to be known. Also, the outer radius (R) and inner radius (r) of the curvature of the module needs to be known. These parameters can be seen in Figure \ref{fig_drawings}. Then, the area occupied by the fringes is given by,
\begin{equation}
\label{equ_A_1}
A_1 = \frac{\pi \left(R^2 - r^2 \right)}{4}
\end{equation} 

This area is equal to the area formed by the triangular fringes. It is given by,
\begin{equation}
\label{equ_A_2}
A_2 = N \times \frac{bh}{2}
\end{equation}

Equating both \ref{equ_A_1} and \ref{equ_A_2}, we get,
\begin{equation}
\label{equ_N}
N = \frac{\pi \times \left(R^2 - r^2 \right)}{2bh}
\end{equation}

This gives the number of fringes (N) obtained using the given parameters as input.


\subsection{Drive Design}
\label{subsec_drive_design}
\begin{figure}[H]
\begin{center}
\includegraphics[scale=0.5]{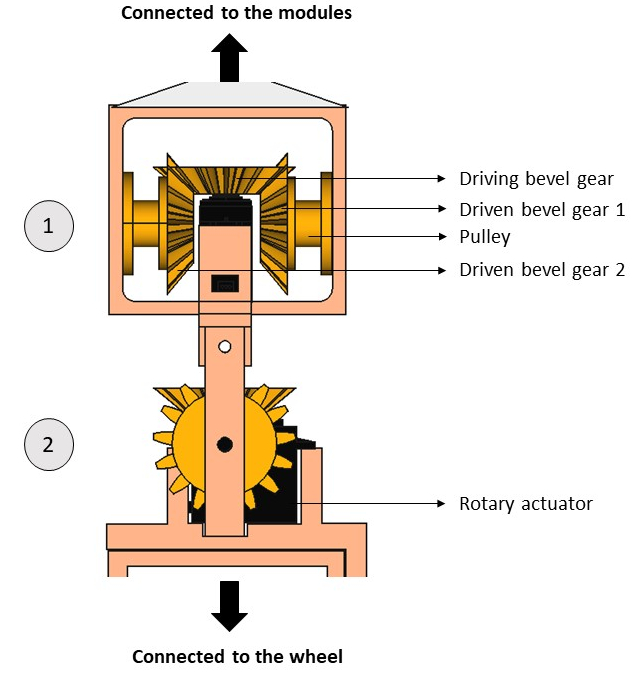}
\caption{The drive design where there are two sets of differential drives. The one end is connected to the modules and the other end is connected to the wheel to convert it to a mobile robot}
\label{fig_drive_design}
\end{center}
\end{figure}

The four modules connected in series are controlled using two differential drives. This is shown in Figure \ref{fig_drive_design}. Each of the differential drives is actuated using a rotary actuator. Therefore two rotary actuators are needed to exhibit pitch and yaw motion by the soft modules of the robot. Lengthening the cable on one side shortens the cable on the other side of the drive. This is because the cables are wound in the same direction. The differential drive has the driving bevel gear connected to two driven bevel gears, 1 and 2. The driven bevel gears have a pulley attached to them. This is to wind the cables that are used to drive the modules. In differential mechanism 1, as the driving gear rotates in a clockwise direction, the bevel gear one also rotates in a clockwise direction, and the bevel gear 2 rotates in an anti-clockwise direction. This makes the modules move in the right direction. If the direction of the driving gear is reversed, the modules move in the yaw left direction.
Similarly, when the driving bevel gear of the differential mechanism 2 rotates in the clockwise direction, the module would pitch down. If the direction of the driving gear is reversed, the module will pitch up. Therefore, the position of the soft modules end effector can be changed by controlling the direction of the driving gears in the differential mechanism. This drive mode of controlling the soft continuum robots is introduced in this paper, and to the best of the authors knowledge, it does not exist in the literature.  

\subsection{Design and Application of Soft Continuum Manipulator}
\label{subsec_scm}
\begin{figure}[H]
\begin{center}
\includegraphics[scale=0.17]{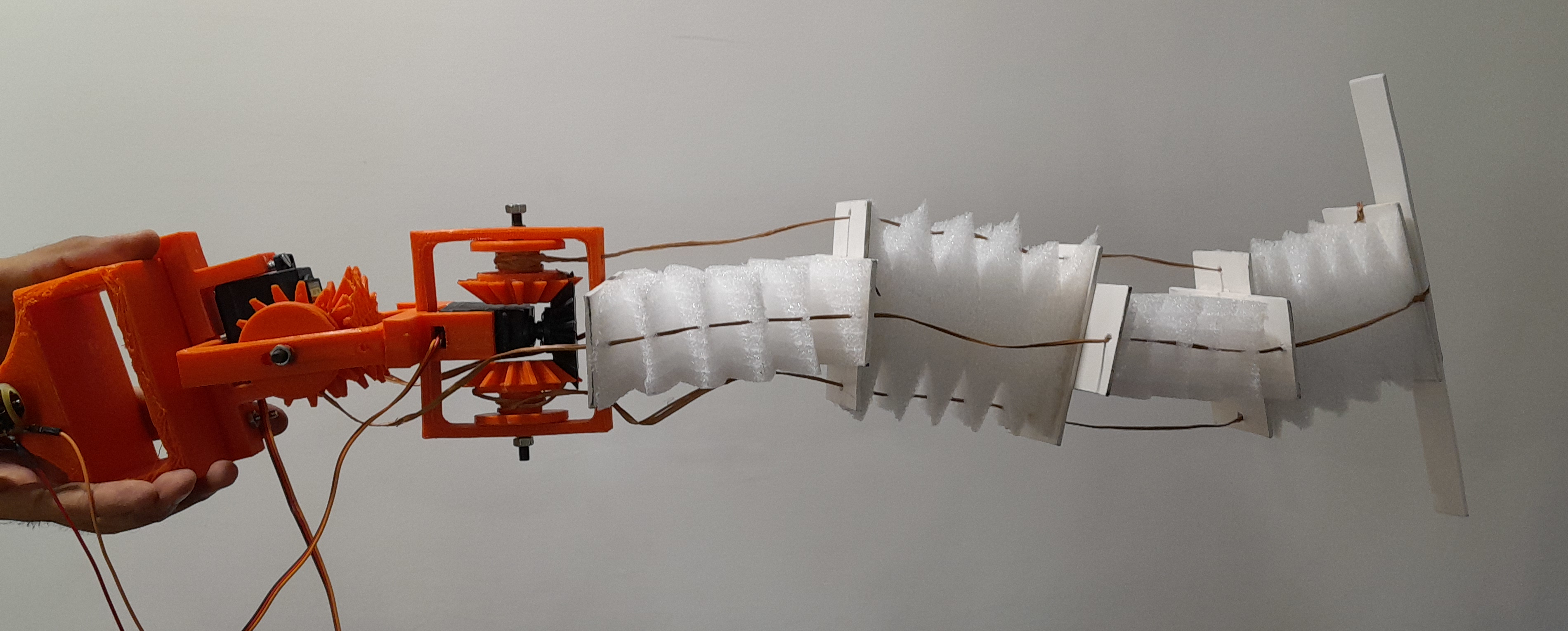}
\caption{The prototype of the soft continuum manipulator where the modules are at the initial position}
\label{fig_prototype}
\end{center}
\end{figure}

The SCM comprises four modules described in Section \ref{subsec_module}. It is driven by the two drives and four cable systems described in Section \ref{subsec_drive_design}. The rotary actuators are continuous rotation servo motors with a stall torque of 16kg/cm at 7.2V. Since the modules and the driving unit are light in weight, the manipulator can be used as a handheld robot. This is shown in Figure \ref{fig_prototype}. In the figure, the modules are in the initial position. It has a thin rectangular strip in the end effector to provide better turning on the floor during the mobile robot mode. We plan to attach a camera to the end effector in the future version. This will be useful for searching inside debris post earthquake or inside collapsed buildings. The rescuer has to insert the modules inside debris. The structure of the module can be changed using the joystick-driven drives. This adjusts the length of the cable for better motion of the modules inside debris. When the void is huge and beyond the reach of the manipulator, the soft robotic arm can be converted to a mobile robot mode and let inside debris for the search operation. Since the SCM is modular, it can be replaced when damaged.   

\subsection{Design and Working of Soft Snake Robot}
\label{subsec_ssr}
The SSR is an extension of the SCM. When a dual shaft two-wheeled DC motor is fitted to the front of the manipulator, the robot becomes an SSR. The structure of the modules needs to be adjusted to create sufficient traction on the floor for the SSR. When the modules are in initial position, the DC motor can be actuated for the motion in the forward or backward direction. To make the robot take a turn left or right, the modules are made to yaw left or right respectively and press on the ground using the rectangular strip at the end effector. Then, the DC motor is actuated, and the robot takes a turn accordingly. The modules turn using the drives, which are controlled manually by the 2-axis joystick. 

\begin{figure}[H]
\begin{center}
\includegraphics[scale=0.2]{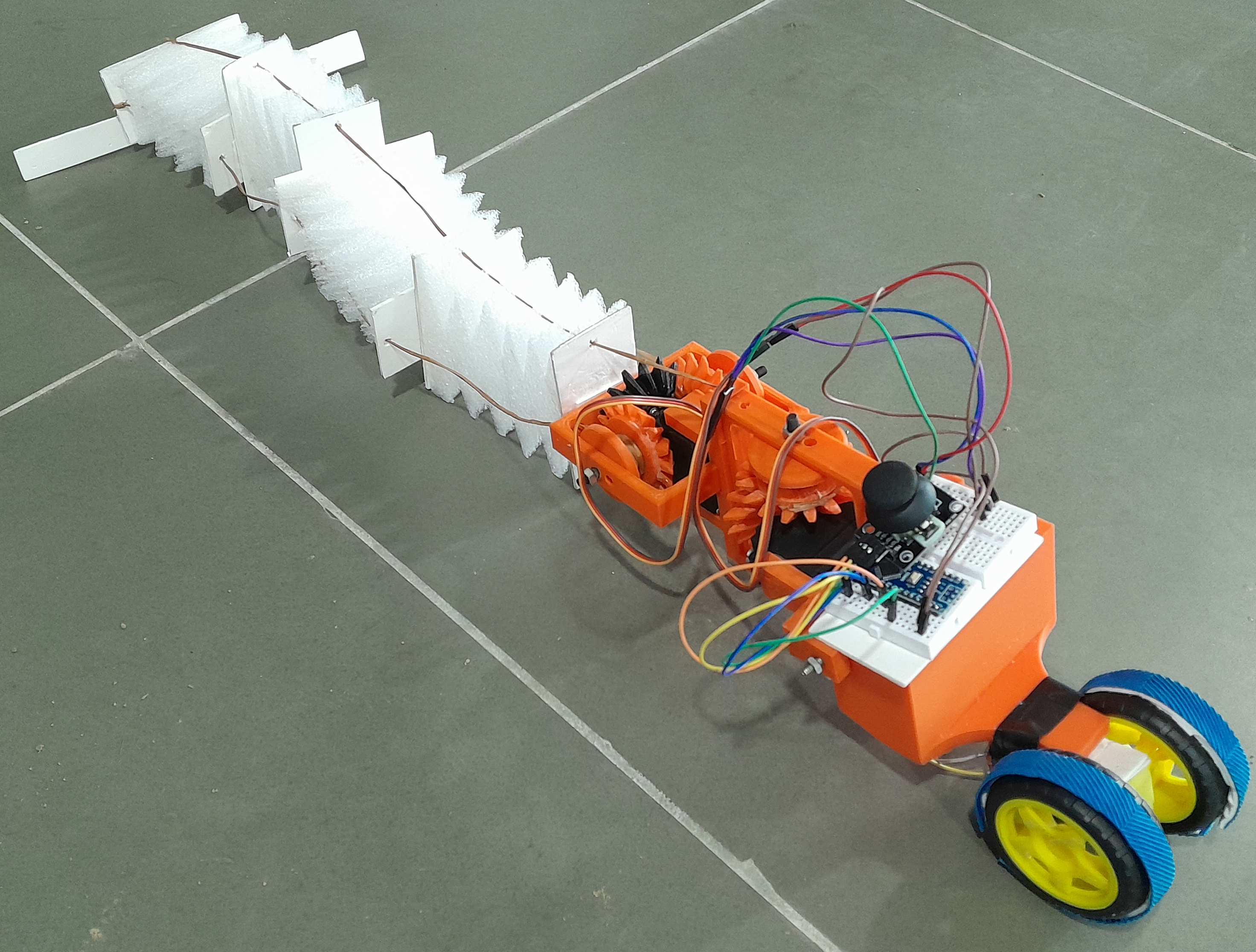}
\caption{The prototype of the soft snake robot}
\label{fig_mobile_robot}
\end{center}
\end{figure}

\subsection{Modeling of the Robotic System}
\label{subsec_model}
The kinematic modeling of the SCM and SSR is done based on the works available in the literature. These will lead to the dynamic analysis in the future. This is a step towards making robots controllable and autonomous in the future. In this section, the results of the kinematic analysis are presented. 

\subsubsection{The soft continuum manipulator}
The kinematic modeling of the SCM was done using the Denavit-Hartenberg (DH) method. The process of using this DH method for deriving kinematics of the continuum manipulators can be referred to in \cite{bamdad2019kinematics, shi2024position}. The DH parameters of the SCM are presented in Table \ref{tab_dh_param}. The kinematic modeling of the continuum robot can also be done by modifying the kinematic equations presented in \cite{zhou2022bioinspired}. It is also helpful to calculate the workspace of the manipulator.     

\begin{table}[H]
\centering
\caption{Denavit-Hartenberg parameters for a 4 module soft continuum manipulator}
\begin{tabular}{|c|c|c|c|c|}
\hline
Joint (i) & \( \theta_i \) & \( d_i \) & \( a_i \) & \( \alpha_i \) \\
\hline
1 & \( \theta_1 \) & 0 & \( l_1 \) & $90^\circ$ \\
2 & \( \theta_2 \) & 0 & \( l_2 \) & $90^\circ$ \\
3 & \( \theta_3 \) & 0 & \( l_3 \) & $90^\circ$ \\
4 & \( \theta_4 \) & 0 & \( l_4 \) & $90^\circ$ \\
\hline
\end{tabular}
\label{tab_dh_param}
\end{table}

\subsubsection{The soft snake robot}
\begin{figure}[H]
\begin{center}
\includegraphics[scale=0.5]{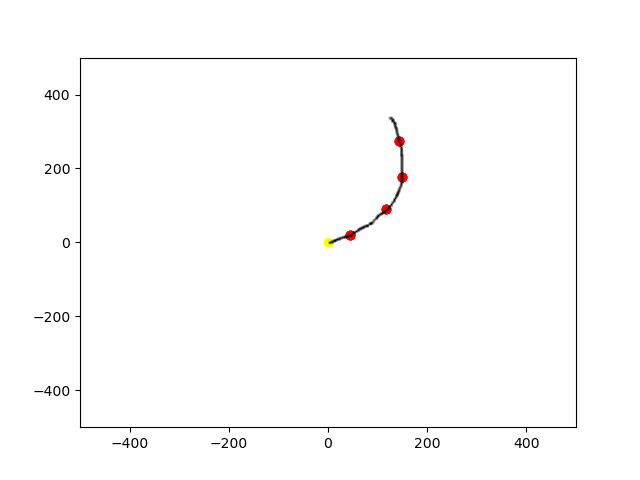}
\caption{The bending of the snake robot in the planar environment where all the angles between successive joints are 25$^o$}
\label{fig_snake_model}
\end{center}
\end{figure}

The SSR was theoretically modeled using the method proposed in \cite{luo2014theoretical}. The algorithm was converted to a Python program described in Section \ref{append_Kinematic}. The input angles were fed to the equations, and the position of the snake robot in a plane was observed. Although the continuum section of the robot is spatial, the section is planar when it acts as an SSR. The Figure \ref{fig_snake_model} shows the bending of the snake robot in the planar environment. 

\subsection{Workspace of the Soft Continuum Manipulator}
\label{subsec_workspace}
\begin{figure}[H]
\begin{center}
\includegraphics[scale=0.4]{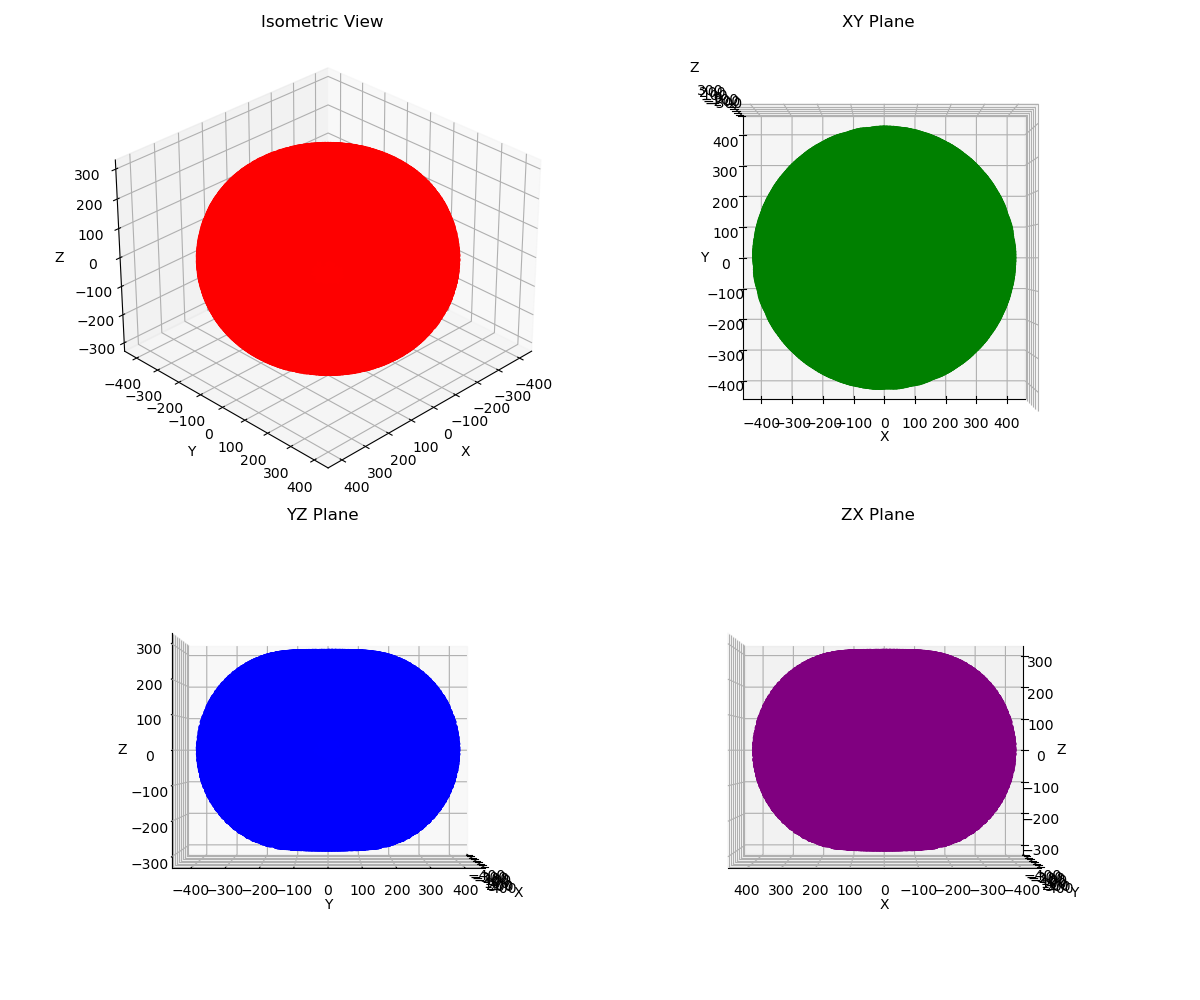}
\caption{The theoretical workspace of the soft continuum manipulator}
\label{fig_workspace}
\end{center}
\end{figure}

Understanding the theoretical workspace is essential to know how far the manipulator's end effector can reach. This work is done by finding the transformation matrix using the information in Table \ref{tab_dh_param} and the method explained in \cite{niku2020introduction}. A Python program was written, and the output obtained is presented in Figure \ref{fig_workspace}. The manipulator's end effector reaches a distance of 400 mm in the X and Y axes and 300 mm in the Z axes. This is experimentally verified in Section \ref{subsec_exp_workspace}. The algorithm is shown in Section \ref{append_workspace}.

\section{Optimal Design of the Module}
\label{sec_optimal_design}
\begin{figure}[H]
\begin{center}
\includegraphics[scale=0.35]{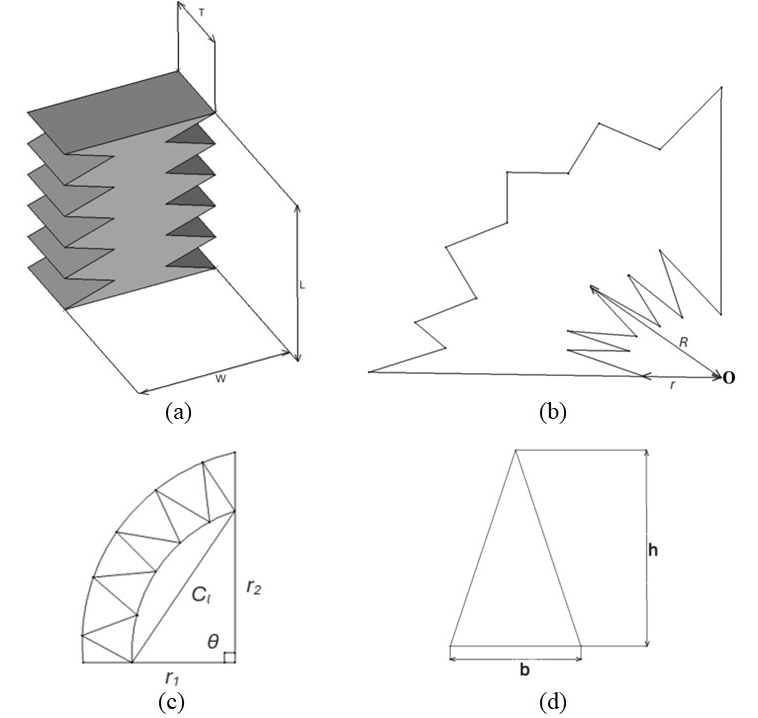}
\caption{(a) The dimensions of the module (b) The module forming 90$^O$ turn (c) The module forming 90$^O$ turn where the radius of curvature is different (d) The dimensions of the triangle that forms the fringes.}
\label{fig_drawings}
\end{center}
\end{figure}

\begin{figure}[H]
\begin{center}
\includegraphics[scale=0.35]{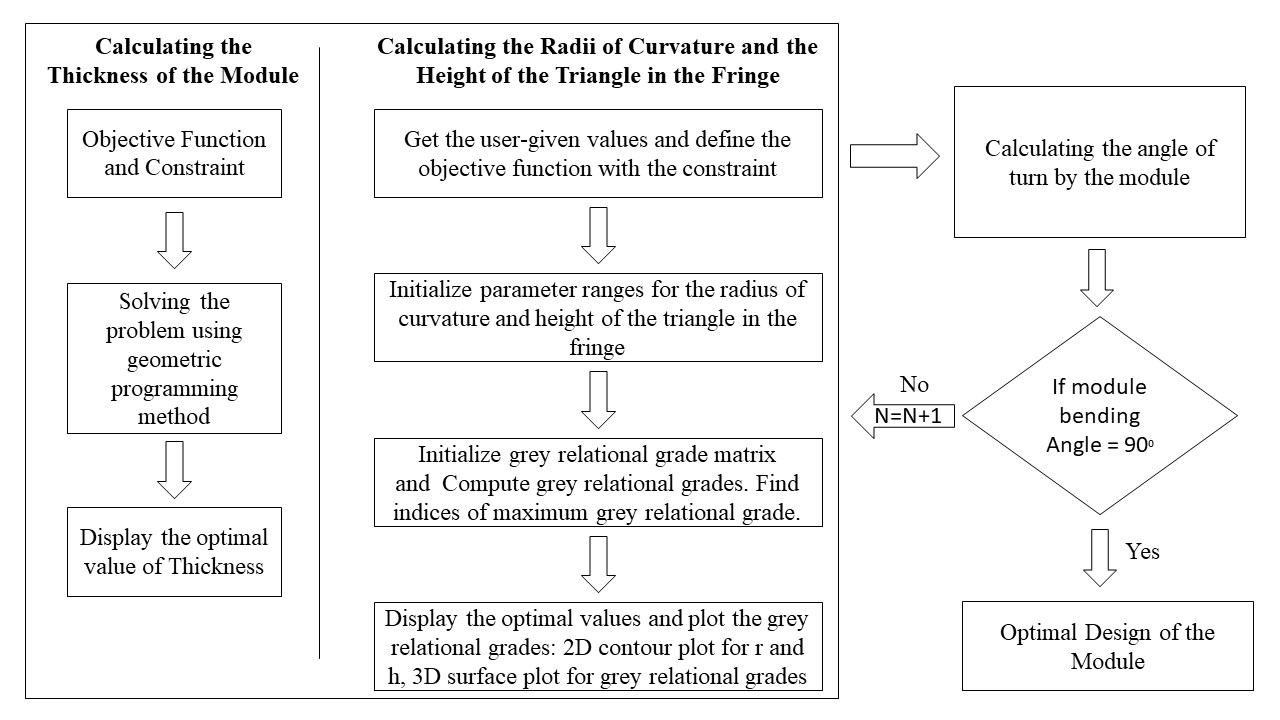}
\caption{The overview of the optimization process to find the dimensions of the module}
\label{fig_opti_overview}
\end{center}
\end{figure}

The dimensions of the module are determined in this section. Figure \ref{fig_drawings} shows the various dimensions of the modules, such as the length of the module (L), the width of the module (W), the thickness of the module (T), base of the triangular fringe (b), height of the triangular fringe (h), and radii of inner (r) and outer curvatures (R). The fringes are mirrored about the length of the module by the middle of the vertical axis. The thickness of the module is calculated using the geometric programming method. Then, the radii of curvatures and height of the fringe are optimized using the Grey Relation analysis. Using these values the angle of bending of the module is calculated. If the bending angle is $90^{o}$, the values obtained are considered optimal. If it is less than $90^{o}$, then the number of fringes is increased by one unit, and the optimization process is repeated. This process is shown in Figure \ref{fig_opti_overview}.

\subsection{Calculating the Thickness of the Module}
The value of the thickness is calculated using the geometric programming method. The length (L) and width (W) of the modules, force (F) applied on the module, density (D) of the module, and the maximum allowable bending stress ($\sigma$) are given as input to the optimization process. The value of W is assigned to be \textit{90 mm} since it is the minimum space obtained to mount the module over the drive system. The drive system was designed to meet the torque required to operate the manipulator modules. The length (L) of the module is fixed to be \textit{100 mm} so that one face of the cuboid (module) is approximately a square. The density of the module is approximately \textit{20 kg/$m^3$}. The maximum allowable bending stress is considered as \textit{160000 N/$m^2$}. The thickness (T) is the variable that is to be calculated. 
The objective function is to minimize: 
\begin{equation}
f(T) = D \times L \times W \times T 
\end{equation}

The constraint is given by,
\begin{equation}
constraint = \frac{(6 \times F \times L)}{(\sigma \times W \times T^2)}  <= 1
\end{equation} 

Using this objective function and constraint equation, the problem of finding the optimal thickness value is formulated and solved using the geometric programming method. The plot of the objective function against the thickness of the module is shown in Figure \ref{fig_geo_prog_opti}. The algorithm for the process is explained in the appendix, which is in Section \ref{append_geo_pro}.
 
\begin{figure}[H]
\begin{center}
\includegraphics[scale=0.35]{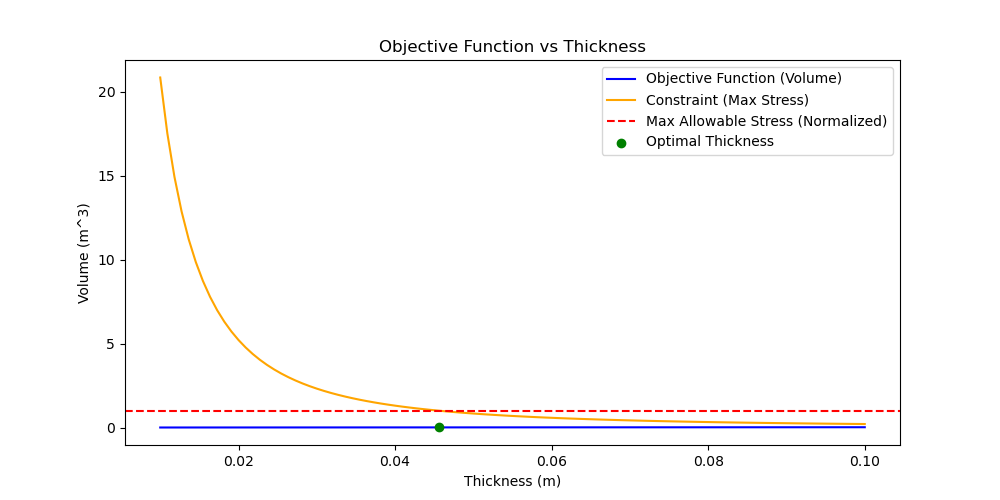}
\caption{The plot showing the optimal thickness value that is plotted against the objective function}
\label{fig_geo_prog_opti}
\end{center}
\end{figure}

\subsection{Calculating the Radii of Curvature and the Height of the Triangular Fringe}
In this stage of optimization, the values of the number of fringes (N) and the base (b) dimensions are fixed. The value of \textit{b} is calculated using the equation \ref{equ_base_fri}. The value of N is started from \textit{1} and increased till the angle of turn ($\theta$) is approximately $90^{o}$. In this work, the value of N = 5 and the b = 20 mm.    

\begin{equation}
\label{equ_base_fri}
b = \frac{L}{N}
\end{equation}

The objective function is given by equation \ref{equ_N}. The value of \textit{R} is given by equation \ref{equ_R}, which is the constraint. 

\begin{equation}
\label{equ_R}
R = r + h
\end{equation}

The initialize parameter ranges for \textit{r} and \textit{h} are given as follows:
 
r = 25 mm to 30 mm

h = 30 mm to 35 mm

\begin{figure}[H]
\begin{center}
\includegraphics[scale=0.35]{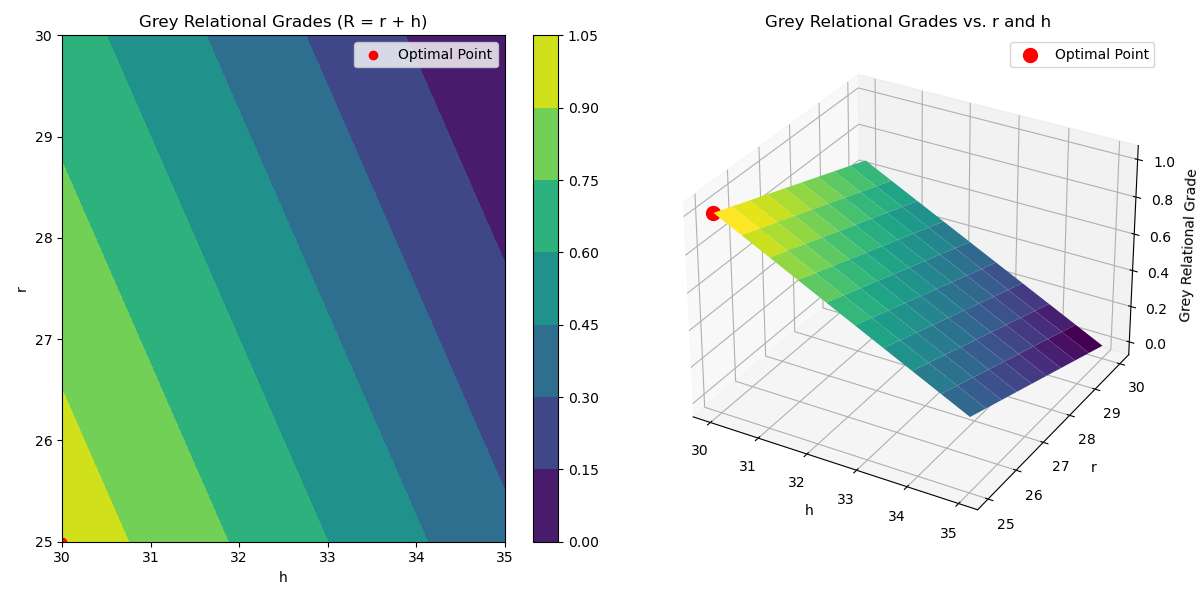}
\caption{The Grey relation grades and the optimal value represented as a red point}
\label{fig_grey_grades}
\end{center}
\end{figure}

The width of the module constrains these ranges. The Grey Relational Grade matrix is formed, from which the Grey Relational Grades are calculated. The indices of the maximum Grey Relational Grade are then identified, which leads to the calculation of optimal values. The optimal \textit{R = 55 mm}, \textit{r = 25 mm} and \textit{h = 30 mm}. Figure \ref{fig_grey_grades} (a) shows the 2D contour plot for r and h. Figure \ref{fig_grey_grades} (b) represents the 3D surface plot for Grey Relational Grades. The optimal point is plotted as a red dot in the graphs. The appendix in Section \ref{append_grey} explains the Grey Relation analysis algorithm.

\subsection{Calculating the angle of turn}
The angle of turn ($\theta$) is the bending of the module about the fringes when the cables are actuated by the drive system. From Figure \ref{fig_drawings} (c), it can be noted that the value of $\theta$ is given using the equation \ref{equ_theta_t} as follows (when $r_1$ = $r_2$):

\begin{equation}
\label{equ_theta_t}
\theta = 2 \times \arcsin \frac{C_l}{2r}
\end{equation}  

When the value of $r_1$ and $r_2$ are different, then the equation \ref{equ_theta_t} becomes, 
\begin{equation}
\label{equ_theta_t_2}
\theta = 2 \times \arcsin \frac{C_l}{r_1 r_2}
\end{equation}  

There are four modules in the SCR. Each is arranged in a $90^{o}$ offset position. The objective is to make the overall manipulator turn $180^{o}$ about the two axes. Hence, the value of $\theta$ is targeted to reach $90^{o}$ about the axis. 

\subsection{Dimensions of the Module}
The geometric programming was used to find the thickness of the module. The radii of curvatures and height of the triangular fringe were found using the Grey Relation analysis. Therefore, in the above sections, we have found the values of all the design parameters of the module. They are presented in Table \ref{tab_dimensions} in a concise form. 

\begin{table}[H]
\centering
\caption{The dimensions of the module parameters}
\begin{tabular}{|c|c|}
\hline
\textbf{Design Parameter}                & \textbf{Value} \\ \hline
Length of the module (L)                 & 100 mm         \\ \hline
Width of the module (W)                  & 90 mm          \\ \hline
Thickness of the module (T)              & 45 mm          \\ \hline
Outer radius of curvature (R)            & 55 mm          \\ \hline
Inner radius of curvature (r)            & 25 mm          \\ \hline
Base of the triangular fringe (b)   & 20 mm          \\ \hline
Height of the triangular fringe (h) & 30 mm          \\ \hline
Number of fringes (N)                    & 5              \\ \hline
\end{tabular}
\label{tab_dimensions}
\end{table}

\section{Fabrication of the Soft Continuum Robot}
\label{sec_fab}
The modules are made using expanded polyethylene foam (EPE), and the drive system is additively manufactured using polylactic acid (PLA). The reason for using EPE for fabricating the module is due to the lightweight nature, flexible and elastic properties, durability, shock absorption, water resistance, thermal insulating properties, easy to work with and low cost. The polylactic acid (PLA) filament was used to fabricate the drive system parts since it has good tensile strength, Youngs modulus, ductile, chemical resistance and low cost.  The fabrication procedure of both the module and the drive system are explained in this section.
\subsection{Module}
\begin{figure}[H]
\begin{center}
\includegraphics[scale=0.35]{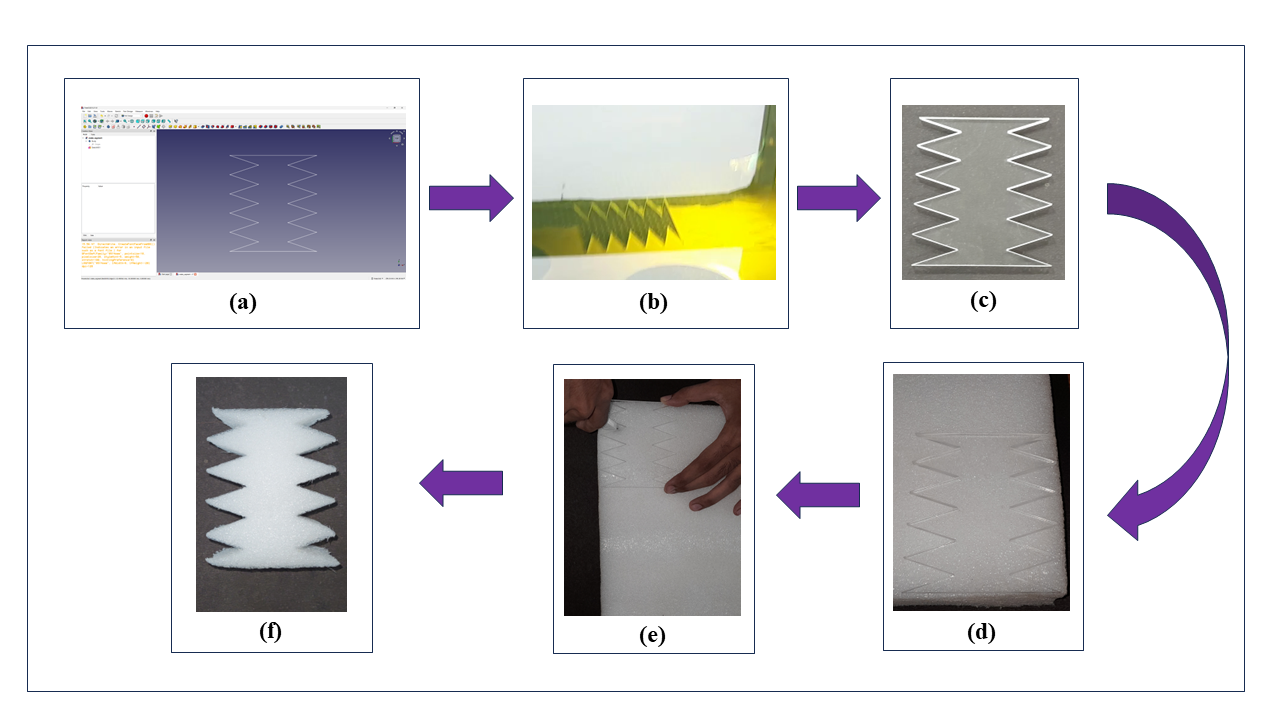}
\caption{The stages involved in the manufacturing of the module (a) Computer-aided drafting of the module design (b) Laser cutting of the template on the acrylic sheet (c) The fabricated acrylic sheet that is to be used as a template (d) Fixing the template on the EPE temporarily (e) Cutting out the modules using a sharp paper cutting knife (f) The fabricated module}
\label{fig_module_manu}
\end{center}
\end{figure}

The 2D cross-section of the module is drawn in the \textit{FreeCAD$^{TM}$} software in the draft mode. This is shown in Figure \ref{fig_module_manu} (a). It is then fed to the laser cutting machine to cut the template on a 10mm thick acrylic sheet. This can be seen in Figure \ref{fig_module_manu} (b). Then, the template of the module made in acrylic material is removed from the laser cutting machine, as shown in Figure \ref{fig_module_manu} (c). The acrylic template of the module is then placed on the EPE and fixed temporarily, as shown in Figure \ref{fig_module_manu} (d). Then, a paper-cutting knife is used to cut out the module manually in the EPE. This process is shown in Figure \ref{fig_module_manu} (e). Finally, the EPE module is removed, and the acrylic template is removed. The final module is shown in Figure \ref{fig_module_manu} (f). Then the holes for the cables to pass through are made using a paper cutting knife by poking it on the fringes. The entire process is shown in Figure \ref{fig_module_manu}. 

\subsection{Drive System}
The gears were modeled using the \textit{FreeCAD$^{TM}$} software in the \textit{Gear} workbench. Then, the pulleys were attached to it. The frames were also designed using the same CAD software. The individual components were exported in the \textit{.stl} (STereoLithography) format. They were imported individually to the \textit{Creality} slicing software where the \textit{G-Code} was generated. The infill density was set at 30$\%$, and the support structure generation was turned on. It was then uploaded to the \textit{Ender 5 Plus} 3D printer. The material used for printing was PLA filament. It took about 36 hours for all the parts to be 3D printed. The bevel gear with pulley required post-processing since the support material was filled inside the hollow pulley region. The rest of the parts did not need much post-processing. Then, the bevel gear without the pulley was attached to the circular horn of the servo motor. The entire setup was then attached, and the cables were connected to the drive system.

\section{Mechatronic design}
\label{sec_mechatronic_design}
The design and fabrication of the robot were explained in the previous sections. In this section, the electronics and control of the SCR are explained. It is challenging to control a soft-bodied robot since it possesses infinite passive degrees of freedom \cite{luo2015slithering}. In our work, we plan to manually control the robot using a simple 2-axis joystick and a two-way switch. 
\subsection{Overall System Architecture}
\begin{figure}[H]
\begin{center}
\includegraphics[scale=0.4]{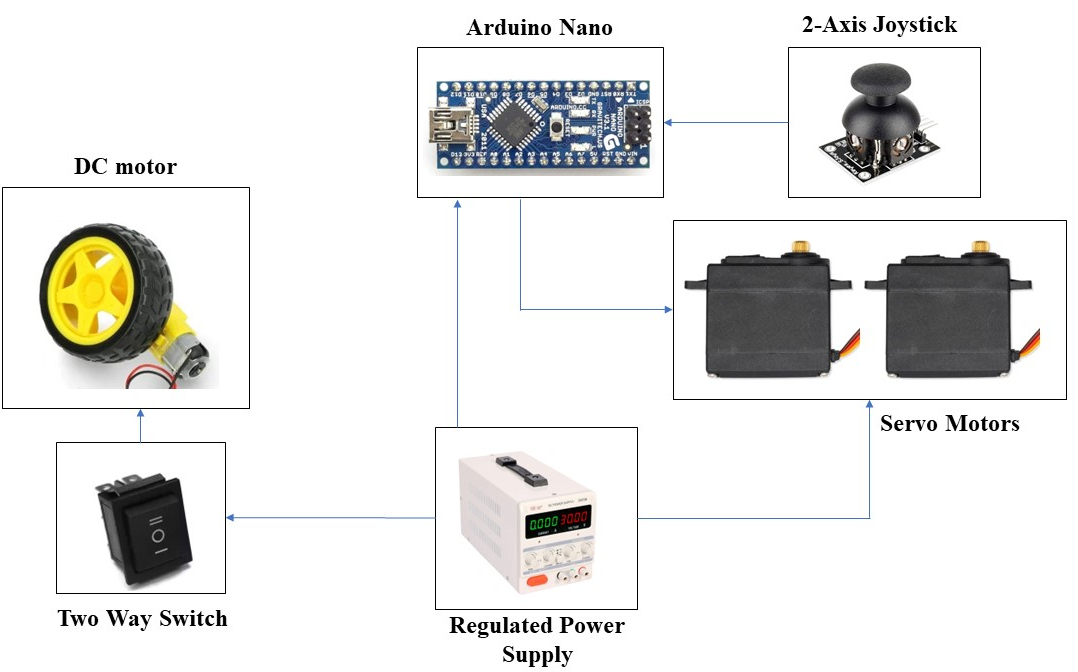}
\caption{The overall block diagram of the electronics involved in the functioning of the robot}
\label{fig_elect_block}
\end{center}
\end{figure}

The overview of the electronics involved in the robot is presented in Figure \ref{fig_elect_block}. The microcontroller used to control the robot is \textit{Arduino Nano}. There is a 2-axis joystick that is used to control the direction of rotation of two servo motors. The two servo motors are of continuous rotation and attached to the drive system. There is a dual-shaft DC motor that controls the motion of the SSR. The direction of rotation of the DC motor is controlled by the two-way switch. The various parameters of the components and their specifications are mentioned in Table \ref{tab_spec}. 
  
\begin{table}[H]
\centering
\caption{The specifications of components used in the SCR}
\begin{tabular}{|l|l|l|}
\hline
\textbf{Component Name}              & \textbf{Parameter} & \textbf{Specification} \\ \hline
\multirow{2}{*}{Dual shaft DC motor} & RPM                & 60                     \\ \cline{2-3} 
                                     & Voltage            & 9V                     \\ \hline
\multirow{2}{*}{Servo motor}         & Stall Torque       & 16 kg/cm               \\ \cline{2-3} 
                                     & Voltage            & 7.4V                   \\ \hline
Arduino Nano                         & V$_{in}$           & 7.4V                   \\ \hline
2-Axis Joystick                      & Voltage            & 5V                     \\ \hline
\end{tabular}
\label{tab_spec}
\end{table}

\subsection{Control of the Soft Continuum Manipulator}
The SCM mode of the robot uses two servo motors, a microcontroller, a 2-axis joystick, and a power supply. The microcontroller is programmed so that when the joystick moves in the up and down direction (say the Y axis), the servo motor of the second differential drive rotates in the clockwise and anti-clockwise direction, respectively. Similarly, when the joystick moves in the right and left direction (the X axis), the servo motor of the first differential drive rotates clockwise and anti-clockwise, respectively. The rest of the functioning of the drives of the SCM was explained in Section \ref{subsec_drive_design}. The servo motors were powered externally, and the control pulses (PWM signals) were given from the microcontroller as per the user input from the 2-axis joystick. The SCM modules would pitch up, pitch down, yaw right, and yaw left, based on the drive being driven by the user using the 2-axis joystick. The experimental validation of the SCM is explained in Section \ref{subsec_expri_mani}.
       
\subsection{Control of the Soft Snake robot}
In the current version of the SSR, the DC motor is not integrated with the microcontroller to simplify the control. When the robot needs to be used in the SSR mode, the dual shaft DC motor must be attached to the front of the electronic chamber. There are two wheels attached to both shafts of the dual-shaft DC motor. These are connected to the bidirectional switch and then to the power supply. When the switch is turned on for clockwise rotation, the SSR moves in the forward direction.
On the other hand, when the motor rotates in the anti-clockwise direction, the robot moves backward. For the robot to take a turn, the tail of the robot is curled out and made stiff about a point by actuating the differential drive system. Then, the robot is moved by actuating the DC motor.  

\section{Working of the Soft Continuum Robot}
\label{sec_work}
The components that were fabricated, as mentioned in Section \ref{sec_fab}, were assembled as shown in Figure \ref{fig_CAD}. The functioning of the SCM for pitch and yaw motion was first verified. Then, the SSR is tested for motion in a straight line and for taking a turn about an axis. These are presented in this section. Then, a few evaluation parameters presented in \cite{bruzzone2012locomotion} were experimentally verified and are presented in Section \ref{sec_work}. 

\subsection{Soft Continuum Manipulator}
\label{subsec_expri_mani}
\begin{figure}[H]
\begin{center}
\includegraphics[scale=0.5]{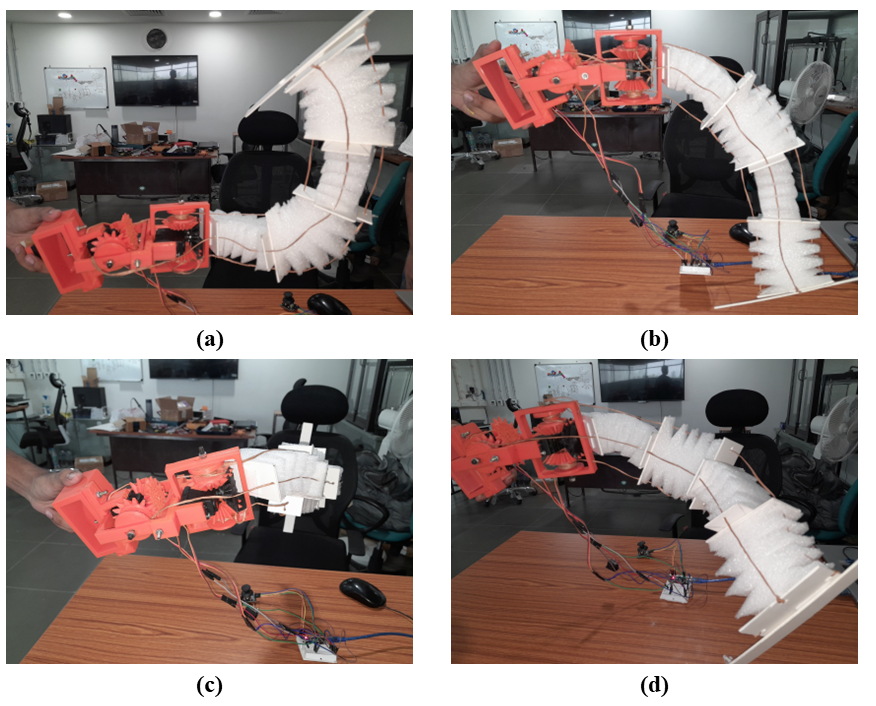}
\caption{The bending of the SCM (a) Pitch up (b) Pitch down (c) Yaw left (d) Yaw right}
\label{fig_experi_mani}
\end{center}
\end{figure}

The SCM is actuated by controlling the 2-axis joystick. The direction of rotation of soft modules when the differential drive is actuated is presented in Section \ref{subsec_drive_design}. The experimental results are shown in Figure \ref{fig_experi_mani}. The pitch up and pitch down are shown in Figure \ref{fig_experi_mani} (a) and (b). The yaw left and yaw right is shown in Figure \ref{fig_experi_mani} (c) and (d). The SCM can also be driven to any other position based on manual commands using the 2-axis joystick. In this version of the robot, pitch or yaw motion can be performed at the same time. It is by holding the joystick at an intermediate position between the axes. In future iterations, two joysticks will be used independently as input devices for pitch and yaw motions. The end-effector will have a camera that will feed live inputs about the condition inside debris to the rescuer, enabling him to make decisions accordingly.  

\subsection{Soft Snake Robot}
In the SSR mode, the SCM is converted into a mobile robot. The SSR can move in a straight line, take a turn, and traverse obstacles. This is done by using the active wheels, driven by the dual shaft DC motor attached to the electronics chamber. In this section, the straight line motion and turning motion of the SSR are explained in detail.  

\subsubsection{Moving in a straight line}
\begin{figure}[H]
\begin{center}
\includegraphics[scale=0.5]{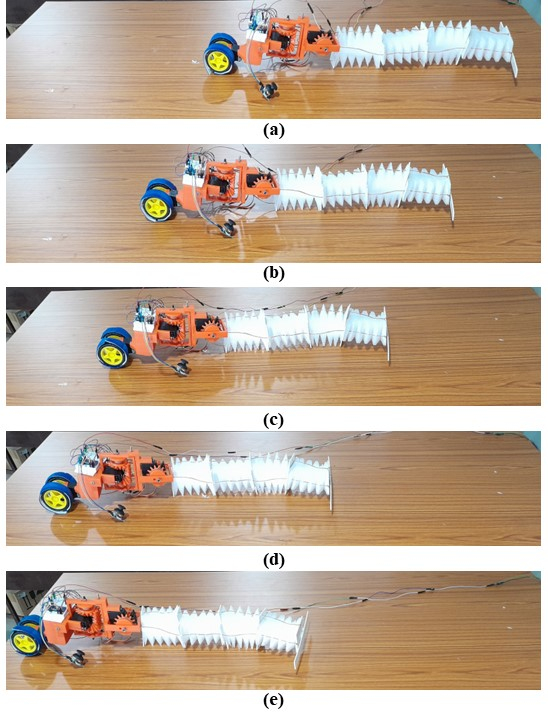}
\caption{The snapshots of soft snake robot mode moving in a straight line}
\label{fig_straight_mobile}
\end{center}
\end{figure}

Moving in a straight line is a joint motion exhibited by most of the robots. Here in this work, we make the SSR move on a straight line by placing the tail of the SCR in the initial position, as shown in Figure \ref{fig_prototype}. This can also be seen in Figure \ref{fig_straight_mobile} (a). By enabling the two-way switch to rotate in the forward direction, the SSR moves in front, as shown in the sequence of images in Figure \ref{fig_straight_mobile} (a)-(e). When the direction of rotation of the wheels are reversed using the switch, the SSR moves in the reverse direction. Thus, the forward and backward motions are possible using SSR.

\subsubsection{Turning motion}
\begin{figure}[H]
\begin{center}
\includegraphics[scale=0.5]{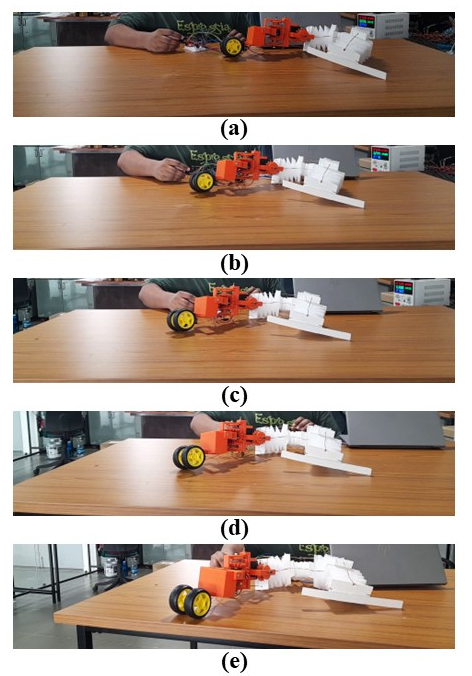}
\caption{The snapshots of soft snake robot mode of the continuum robot taking a left turn}
\label{fig_turn}
\end{center}
\end{figure}

The turning motion is a necessary action that needs to be exhibited by the SSR. The robot can explore a wider area when it turns and traverses on uneven terrains. The tail exhibits yaw left or yaw right motion for the SSR to turn. Then, the rectangular strip at the end effector presses on the terrain for better friction. Then, the DC motor is made to move in the forward or backward direction, and the SSR makes a turn. Figure \ref{fig_turn} shows the SSR taking a left turn. This is done by performing the yaw right motion, shown in Figure \ref{fig_turn} (a). The rest of the turning motion can be seen in Figure \ref{fig_turn} (b)-(e). Thus, the SSR can turn, although it has only one set of wheels driven by a DC motor.

\section{Experiments with the Soft Continuum Robot}
\label{sec_experi}
\subsection{Soft Continuum Manipulator}
The deformation of the module is simulated and presented in this section. The workspace dimensions are also verified experimentally. These are as follows:

\subsubsection{Deformation of the Module}
An FEA software was used to perform a static structural analysis to simulate the deflection while bending towards the right through Finite Element Analysis (FEA) on one of the soft segments of the SCM. 

\begin{figure}[H]
\begin{center}
\includegraphics[scale=0.4]{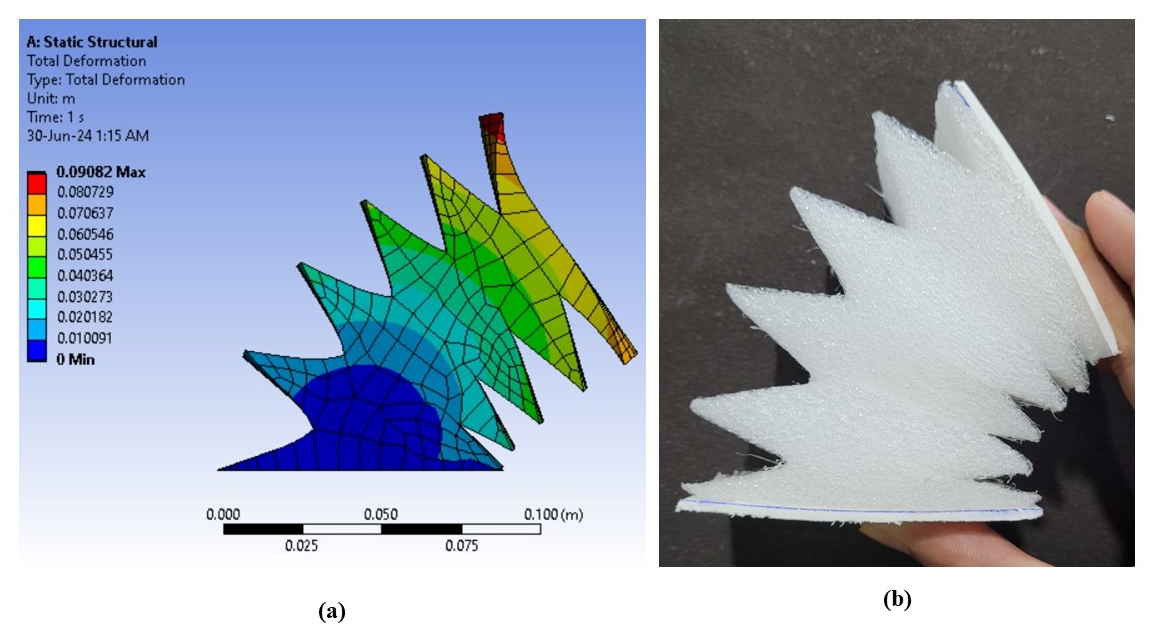}
\caption{The finite element analysis of the module (a) Deformation of the module in the FEA software (b) Deformation of the module in the experiment}
\label{fig_fea}
\end{center}
\end{figure}

The stages involved in the analysis are: (1) Extract the stress-strain data from the reference curve and load to the FEA software. (2) Create a soft-segment CAD model in FEA software. (3) Fix the base of the segment by applying a fixed boundary constraint. (4) Apply the relevant forces and moment. (5) Create a mesh of the model for the analysis (6) Run the simulation. (7) Extract the results (Figure \ref{fig_fea} (a)) and compare them with the physical model (Figure \ref{fig_fea} (b)).

The stress-strain curvature of the material: The properties of Expanded Polyethylene (EPE) \cite{liu2019towards} is loaded into FEA software to model the non-linear properties properly. Subsequently, the base of the model is given a fixed constraint; then, a moment of 300N-m and a downward vertical force of 50N is applied to the top-right edge of the model to simulate the cable pulling at that location. From the displacement diagram (Figure \ref{fig_fea} (a)) the applied loads are enough to produce a relative angle of approximately 80$^O$ between the top and bottom face.  

The high torque (300Nm) and forces (50Nm) are required to produce the quarter circle shape with maximum deformation at the top, approaching 0.1m, while the bottom (fixed constraint) remains stationary. However, the motor used in the physical model can only supply a torque of 16 kg cm, enough to shape the model as a quarter-circle and a semi-circle circle shape when the four soft segments are connected in series. Thus, showcasing the non-linear behavior, which is advantageous to us. This implies more number of soft segments yield lower actuation forces.

\subsubsection{Verification of the Workspace Dimensions}
\label{subsec_exp_workspace}
\begin{figure}[H]
\begin{center}
\includegraphics[scale=0.4]{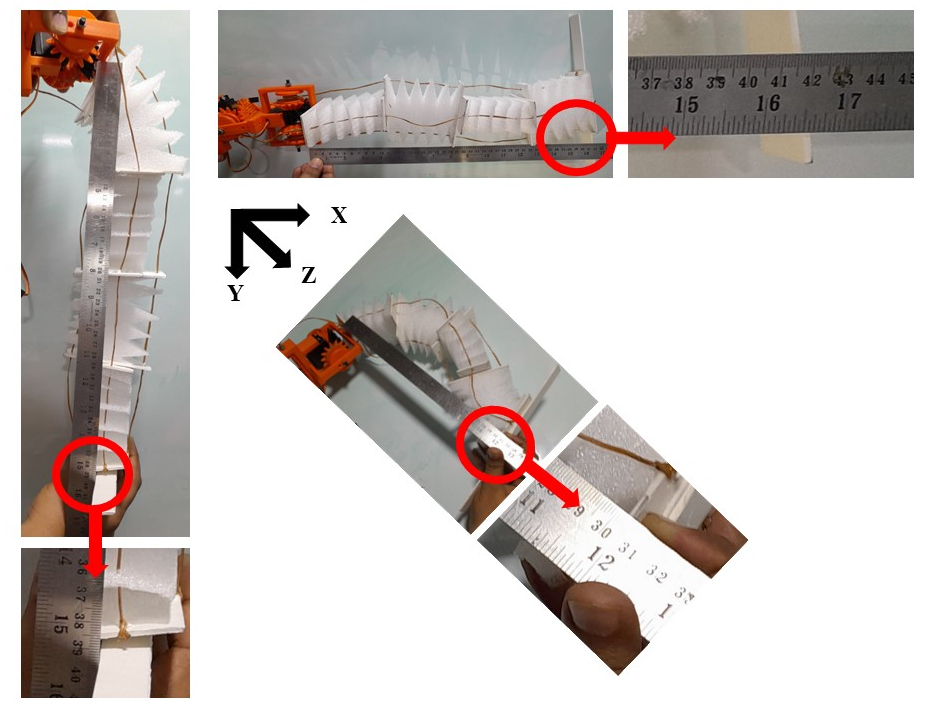}
\caption{The measurements of the endpoints in the workspace for the physical prototype}
\label{fig_exp_workspace}
\end{center}
\end{figure}

The workspace of the SCM is measured by bending the modules about the X, Y, and Z axis. As predicted in Section \ref{subsec_workspace}, the measured values are 400 mm about the X and Y axis and 300 mm about the Z axis. These measurements were physically done and shown in Figure \ref{fig_exp_workspace}. 

\subsection{Soft Snake Robot}
\begin{figure}[H]
\begin{center}
\includegraphics[scale=0.4]{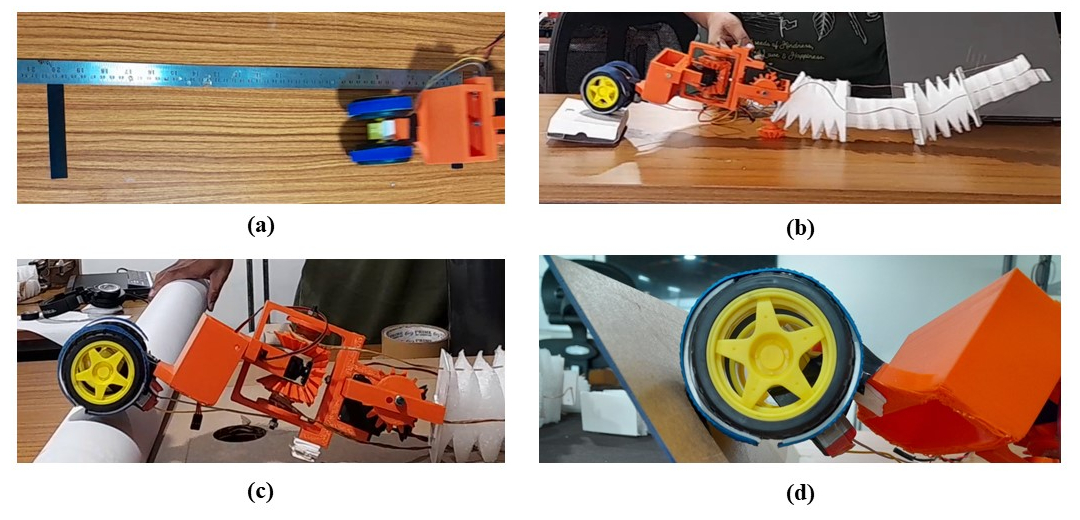}
\caption{The snapshots of soft snake robot (a) Travelling at maximum speed (b) Crossing an obstacle (c) Crossing a step (d) Crossing a slope}
\label{fig_evalu}
\end{center}
\end{figure}

The evaluation of the snake robot is done in this section. It is done using the evaluation parameters that are given in the works of \cite{bruzzone2012locomotion}. The features considered are maximum speed, ability to cross obstacles, ability to climb a step, and ability to climb a slope. The corresponding evaluation parameters are experimented with to find the rank. Finally, based on rank, the robot is classified as a low, medium, or high-fidelity robot. The evaluation of the robot is shown in Figure \ref{fig_evalu}.

\subsubsection{Maximum Speed}
The maximum speed of the SSR is evaluated by moving the robot between two points that are 0.5 m away. The time taken to cross is noted using a stopwatch. Then, the maximum speed is calculated by finding the ratio between the maximum speed of the robot and the total length of the robot. It was experimentally found that the robot traveled at 0.37 m/s. The length of the robot was 0.7 m. The ratio is 0.5285/s, which falls under the medium range as mentioned in \cite{bruzzone2012locomotion}. Figure \ref{fig_evalu} (a) shows the experimental setup for calculating the maximum speed.

\subsubsection{Crossing a Step or a Stair}
The SCR was designed with the ability to cross debris since it is commonly found during search operations in collapsed buildings post earthquake. This version of SCR is capable of crossing small steps. The SSR was made to move in a straight line and then traverse over an obstacle that was 30 mm in height. The radius of the wheel was measured to be 35 mm. This can be seen in Figure \ref{fig_evalu} (b). In the future, attempts will be made to make the SSR climb objects 2-5 times its height. This will be done by pressing the tail on the ground and using more powerful DC motors. Thus, it has been experimentally shown that the SSR is capable of traversing obstacles that are as high as the radius of the wheel. 

\subsubsection{Crossing an Obstacle} 
The SSR was evaluated to cross an obstacle with a semi-circular lateral profile. The robot was able to climb a profile that had a radius of 55 mm. This parameter of climbing a profile is calculated by finding the ratio between the maximum height of the obstacle and the height of the robot. The height of the robot is 120 mm. Therefore, the ratio is 0.4583. This makes the robot fall under the medium range category as mentioned in \cite{bruzzone2012locomotion}. The experimental validation is demonstrated in Figure \ref{fig_evalu} (c).

\subsubsection{Climbing a Slope}
The ability to climb slopes is an essential feature of a robot. In this work, the SSR was made to climb slopes 15$^o$ to 75$^o$. It was found that the robot could climb the slope with a maximum inclination of 66$^o$. This feature makes the SSR fall in the higher range in the slope climbing ability as mentioned in \cite{bruzzone2012locomotion}. The experimental validation is shown in Figure \ref{fig_evalu} (d).

\section{Discussion}
\label{sec_discussion}
This work presented the design, fabrication, and workings of the dual soft continuum robot. This robot can be used as a soft continuum manipulator and a soft snake robot. The SCM can be fitted with a camera and inserted inside the debris during  post earthquake search operation. When the void is suitable for the SSR to pass through, the wheel can be attached, and the robot can be used as an SSR. This SSR can pass inside the debris where humans can not reach and can search for survivors. It would be able to interact with its environment due to the combination of rigid and soft parts; the soft manipulator can be used to clear the pathway from debris and assist in its locomotion, like pushing the head up while climbing. Their soft nature is safer for human interaction.

The workspace of the SCM was calculated in Section \ref{subsec_workspace}. It was experimentally verified in Section \ref{subsec_exp_workspace}. This shows that the manipulator can reach a maximum of 400 mm in the X and Y axes and 300 mm in the Z axes. The modules are soft, and hence flexible. This enables the manipulator to reach and observe the condition inside the debris even if the trajectory is complex; that is, it can conform to complex shapes, making hard-to-reach areas accessible. The manipulator is also light in weight. The drive system has only two motors to operate the manipulator. Each motor can turn the manipulator in the positive and negative directions of the yaw and pitch axis. Therefore, we can control the manipulator by using only two motors instead of four. This aids in easy transportation and handling during operation. The soft modules can be easily manufactured once the acrylic template is available. Therefore, the modules can be replaced if they are worn out. The fabrication of the module is simple and can be quickly done with a few attempts of practice. When comparing this SCM with the other soft continuum manipulators, it has a rectangular cross-section that provides preferential bending towards the four main directions. It also includes resistance towards bending because of the greater area moment of inertia. It can be made with materials that are readily available worldwide.  

The SSR should be capable of traversing debris to reach areas inside the collapsed building. There were several evaluation criteria to measure the performance of the robot, presented in \cite{bruzzone2012locomotion}. Table \ref{tab_evalu} presents a few related to the SCR. The maximum speed criteria classifies the robot to be a medium-range robot. The reason why it did not fall under high-range robots could be because the length of the robot is much greater \textit{(700 mm} compared to the height of the robot \textit{(120 mm)}. This causes the robot to drag. The four modules need to be curled up to reach higher speeds to prevent this. The step climbing ability has a value of \textit{0.25}, making the robot fall under the medium range robot. This could be because of the high speed and low torque of the DC motor, which prevents it from climbing higher steps. The obstacle crossing capability has a value of \textit{0.4583} and makes the robot a medium-range robot. The reason for this again could be the high speed and low torque of the DC motor. The slope climbing ability is assessed to have a value of \textit{66$^o$}, classifying it as a high-range robot. This is because the robot is longer than the height, which makes it climb steep slopes. 

Thus, this dual robot can be used as a manipulator during the initial stages of the search. In the later stages, it can be attached to a DC motor, which makes it an SSR. This robot has a broad workspace and a medium-high performance as a mobile robot. This robot maintains the advantages of a soft robot as it can still fit and conform to hard-to-reach areas.

\begin{table}[]
\centering
\caption{Evaluation criteria}
\begin{tabular}{|l|l|l|l|}
\hline
\multicolumn{1}{|c|}{\textbf{Feature}}                                  & \multicolumn{1}{c|}{\textbf{Criteria}}                                                                                                                                         & \multicolumn{1}{c|}{\textbf{Value}} & \multicolumn{1}{c|}{\textbf{Inference}} \\ \hline
Maximum speed                                                           & \begin{tabular}[c]{@{}l@{}}Find the ratio between \\ the maximum speed \\ achieved by the robot \\ to the overall length \\ of the robot.\end{tabular}                         & 0.5285 s$^{-1}$                     & Medium Range                            \\ \hline
Step/Stair climbing                                                     & \begin{tabular}[c]{@{}l@{}}Find the ratio between \\ the maximum height \\ of a step that the robot \\ can cross to the height \\ of the robot.\end{tabular}                   & 0.25                                & Medium Range                            \\ \hline
\begin{tabular}[c]{@{}l@{}}Obstacle crossing \\ capability\end{tabular} & \begin{tabular}[c]{@{}l@{}}Find the ratio between \\ the maximum height \\ of a semi-circular obstacle \\ that the robot can cross \\ to the height of the robot.\end{tabular} & 0.4583                              & Medium Range                            \\ \hline
\begin{tabular}[c]{@{}l@{}}Slope climbing \\ capability\end{tabular}    & \begin{tabular}[c]{@{}l@{}}Find the maximum \\ inclination of a slope \\ with friction coefficient \\ greater than 0.5 that \\ the robot can climb.\end{tabular}               & 66$^o$                              & High Range                              \\ \hline
\end{tabular}
\label{tab_evalu}
\end{table}

\section{Limitations and Future directions}
\label{sec_lim_fut_dir}
In this section, the limitations of the robot that is presented in this work are presented. Also, the future direction for the robot design and implementation in real-life scenarios is presented. 
\subsection{Limitations}
The current versions have several limitations, making it a medium-range robot. The chosen DC motor was not powerful enough to make it a high range in the maximum speed feature. For the same reason, the step/stair climbing and obstacle crossing capabilities are inferred to be in the medium range. The assembly of the drives had issues with tolerances, and hence, there was a backlash in the bevel gears during the assembly. There are no sensors in the current version since we focused on the mechanical design of the robot. Since the drive motors are present at one end of the robot, the speed of operation of the SCM is lower than that of the rigid manipulators, where there are actuators at every joint. 

\subsection{Future directions}
The next version of the SCR will be able to interact with humans in two aspects of post-earthquake search operations. The first would be the multi-modal interaction between the rescuer and the SCR. This will be done in such a way that the temporary impairment of the rescuer will not affect controlling the robot. For example, in the post-earthquake search operation in Mexico (in the year 2017), an incident of creating shade (by a human) over the laptop monitor was needed to view the camera stream from the snake robot \cite{whitman2018snake}. During such situations, multi-modal interaction would be of greater use. The second interaction is between the trapped survivor and the robot. This should be possible by fitting suitable sensors to the end effector of the SCM. Then, the DC motor used in the current version will be replaced by a powerful motor that will climb steeper stairs and steps. Also, the distance between the two wheels will be increased for better robot stability. After studying the environments in the literature, sensors will be fitted to sense the post-earthquake environment. The end-effector of the SCM will be equipped with a camera that can be used to give live feed to the user.

\section{Conclusion}
\label{sec_conc}
The search operation inside the debris due to earthquakes must be done faster. In this work, we presented a dual soft continuum robot that can be used as a soft continuum manipulator and a soft snake robot. The design, fabrication, working, and evaluation of the manipulator and snake robot were detailed. The pitch and yaw motion of the soft manipulator were experimentally demonstrated. The soft snake robot was made to move along a straight line, take a turn, and traverse an obstacle experimentally. Thus, the dual soft continuum robot was successfully designed and experimented with. This robot is ranked as a medium-range robot as per the evaluation criteria considered in the literature. Future works will involve mounting a camera at the end effector of the soft manipulator and using a more powerful DC motor to traverse objects of heights higher than the wheel's height. Also, the soft continuum manipulator would be efficiently used to aid the motion of the DC motors in the soft snake robot mode. 

\section*{References}

\bibliography{mybibfile}

\section{Appendix: Algorithm for Simplified Kinematic Modelling of the SCR}
\label{append_Kinematic}

\begin{algorithm}[H]
\caption{Transformation and Plotting of Linkages}
\begin{algorithmic}
    \State Initialize $count \gets 1$
    \State Set $\theta_i \gets [\frac{25\pi}{180}, \frac{25\pi}{180}, \frac{25\pi}{180}, \frac{25\pi}{180}]$  \Comment{Convert degrees to radians}
    \State Set $l_1, l_2, l_3, l_4 \gets 100$
    \State Compute $\sin(\theta)$ and $\cos(\theta)$ for each $\theta_i$
    \State Define matrices $A \gets [[1, 1, 0, 0]]$ and $D \gets [[1, -1, 0, 0]]$
    
    \State Calculate $x_i \gets [l_1 \cdot \cos(\theta_i[0]), l_2 \cdot \cos(\theta_i[0] + \theta_i[1]), l_3 \cdot \cos(\theta_i[0] + \theta_i[1] + \theta_i[2]), l_4 \cdot \cos(\theta_i[0] + \theta_i[1] + \theta_i[2] + \theta_i[3])]$
    \State Calculate $y_i \gets [l_1 \cdot \sin(\theta_i[0]), l_2 \cdot \sin(\theta_i[0] + \theta_i[1]), l_3 \cdot \sin(\theta_i[0] + \theta_i[1] + \theta_i[2]), l_4 \cdot \sin(\theta_i[0] + \theta_i[1] + \theta_i[2] + \theta_i[3])]$

    \State Compute $X, Y$ as lists of $x_i$ and $y_i$ values
    \State Calculate the pseudoinverse $Z$ of $D$
    \State Define $e \gets [1, 1, 1, 1]$, $N \gets 4$, $m \gets 1$
    \State Calculate angles $\theta_a, \theta_b, \theta_c, \theta_d$ using $\tan(\theta) = \frac{y_i - y_{i-1}}{x_i - x_{i-1}}$
    \State Calculate curvature terms $k_i = [\frac{\theta_i - \theta_{i-1}}{l_1}, \frac{\theta_i - \theta_{i-1}}{l_2}, \frac{\theta_i - \theta_{i-1}}{l_3}, \frac{\theta_i - \theta_{i-1}}{l_4}]$
    \State Compute $p_x, p_y$ as lists of averaged positions
    \State Compute $X_1, Y_1$ as transposed lists of $x_i, y_i$
    \State Calculate $X_{\text{center\_curved}}, Y_{\text{center\_curved}}$ using curvature adjustments
    \State Calculate $X_{\text{center}}, Y_{\text{center}}$ as adjusted positions
    
    \State \textbf{for} $i$ in range($N$):
    \begin{itemize}
        \item Plot center positions $(2 \cdot X_{\text{center}}[i], 2 \cdot Y_{\text{center}}[i])$ as red dots
    \end{itemize}
    
    \State \textbf{for} $i$ in range($N-1$):
    \begin{itemize}
        \item Plot center positions $(3 \cdot (X_{\text{center}}[i] + X_{\text{center}}[i+1]), 3 \cdot (Y_{\text{center}}[i] + Y_{\text{center}}[i+1]))$ as red dots
    \end{itemize}
    
    \State Plot origin $(0, 0)$ as a yellow dot
    \State Save the plot as an image named $count$.png
\end{algorithmic}
\end{algorithm}

\section{Appendix: Algorithm to find the Workspace of the SCM}
\label{append_workspace}

\begin{algorithm}[H]
\caption{Transformation Matrix Calculation and Plotting the Workspace}
\begin{algorithmic}
    \State Initialize $\theta_{\text{real\_1}}$, $\theta_{\text{real\_2}}$, $\theta_{\text{real\_3}}$, $\theta_{\text{real\_4}}$ as arrays from 0 to 90 with 20 steps
    \State Set $\alpha_1$, $\alpha_2$, $\alpha_3$, $\alpha_4$ to 90
    \State Set $a_1$, $a_2$, $a_3$, $a_4$ to 100
    \State Set $d_1$, $d_2$, $d_3$, $d_4$ to 0
    \State Initialize empty lists for $final\_x$, $final\_y$, $final\_z$
    
    \For{each $\theta_1$ in $\theta_{\text{real\_1}}$}
        \For{each $\theta_2$ in $\theta_{\text{real\_2}}$}
            \For{each $\theta_3$ in $\theta_{\text{real\_3}}$}
                \For{each $\theta_4$ in $\theta_{\text{real\_4}}$}
                    \State Compute $matrix_1$, $matrix_2$, $matrix_3$, $matrix_4$ using the given formula
                    \State Compute final transformation matrix $T = matrix_1 \times matrix_2 \times matrix_3 \times matrix_4$
                    \State Append $T[0, 3]$ to $final\_x$, $T[1, 3]$ to $final\_y$, $T[2, 3]$ to $final\_z$
                \EndFor
            \EndFor
        \EndFor
    \EndFor

    \State Initialize 2x2 subplot grid with 3D projection
    \State Plot $final\_x$, $final\_y$, $final\_z$ in isometric view, XY plane, YZ plane, and ZX plane
    \State Adjust layout and save plot as 'transformation\_views.png'
    \State Display the plot

\end{algorithmic}
\end{algorithm}

\section{Appendix: Algorithm for Geometric Modelling}
\label{append_geo_pro}
\begin{algorithm}[H]
\caption{Optimization of Beam Thickness using Geometric Programming}
\begin{algorithmic}[1]
    \State \textbf{Input:} Given parameters
    \State $L = 0.1$ meters \Comment{Length of the beam}
    \State $W = 0.09$ meters \Comment{Width of the beam}
    \State $F = 50$ Newtons \Comment{Applied force}
    \State $D = 20$ kg/m\textsuperscript{3} \Comment{Density of the material}
    \State $\text{max\_stress} = 160000$ N/m\textsuperscript{2} \Comment{Maximum allowable bending stress}
    \State 
    \State \textbf{Initialize:}
    \State Define $T$ as a positive variable \Comment{Thickness of the beam}
    \State 
    \State \textbf{Objective Function:}
    \State Minimize $D \cdot L \cdot W \cdot T$ \Comment{Volume of the beam}
    \State 
    \State \textbf{Constraint:}
    \State $\frac{6 \cdot F \cdot L}{W \cdot T^2} \leq \text{max\_stress}$ \Comment{Maximum bending stress constraint}
    \State 
    \State \textbf{Formulate the Problem:}
    \State Define the optimization problem using geometric programming
    \State 
    \State \textbf{Solve the Problem:}
    \State Use the \texttt{cvxpy} library to solve the optimization problem
    \State 
    \State \textbf{Output:}
    \State Display the optimal thickness $T$ obtained from the solution
\end{algorithmic}
\end{algorithm}

\section{Appendix: Algorithm for Grey Relational Analysis}
\label{append_grey}

\begin{algorithm}[H]
\caption{Grey Relational Analysis}\label{alg:grey_relational}
\begin{algorithmic}[1]
\State \textbf{Input:} $N$, $b$, $r_{\text{range}}$, $h_{\text{range}}$
\State \textbf{Output:} Optimal $r$, $h$, $R$

\State Initialize grey relational grade matrix $grey\_grades$
\For{$i \in \text{range(len(}r_{\text{range}}\text{))}$}
    \For{$j \in \text{range(len(}h_{\text{range}}\text{))}$}
        \State Compute $R = r + h$
        \State Compute $fitness$ using $objective\_function(r, R, h, b, N)$
        \State Compute $min\_value$ and $max\_value$ across $r_{\text{range}}$ and $h_{\text{range}}$
        \State Compute $grey\_grades[i, j] = \frac{max\_value - fitness}{max\_value - min\_value}$
    \EndFor
\EndFor

\State Find indices $(i, j)$ of maximum $grey\_grades$
\State $optimal\_r = r\_range[i]$, $optimal\_h = h\_range[j]$, $optimal\_R = optimal\_r + optimal\_h$

\State \textbf{Plotting:}
\State Plot grey relational grades using contour and 3D surface plots

\State \textbf{Output:}
\State Print optimal $r$, $R$, $h$

\end{algorithmic}
\end{algorithm}

\end{document}